\documentclass{article}

\usepackage{microtype}
\usepackage{graphicx}
\usepackage{subfigure}
\usepackage{booktabs} 

\usepackage{hyperref}

\usepackage[accepted]{icml2024}

\usepackage{amsmath}
\usepackage{amssymb}
\usepackage{mathtools}
\usepackage{amsthm}

\usepackage{multirow}
\usepackage{comment} 
\usepackage{enumitem}

\usepackage{times}
\usepackage{epsfig}
\usepackage{graphicx}

\usepackage[bb=dsserif]{mathalpha}
\usepackage{bm}

\usepackage{booktabs,colortbl,tabularx}
\usepackage{pifont}%

\definecolor{Gray}{gray}{0.9}

\newcommand{\cmark}{\ding{51}}%
\newcommand{\xmark}{\ding{55}}%

\definecolor{battleshipgrey}{rgb}{0.52, 0.52, 0.51}

\usepackage[capitalize,noabbrev]{cleveref}

\theoremstyle{plain}

\theoremstyle{definition}

\theoremstyle{remark}

\icmltitlerunning{Multi-scale Multi-instance Visual Sound Localization and Segmentation}

\begin{document}

\twocolumn[
\icmltitle{Multi-scale Multi-instance Visual Sound Localization and Segmentation}



\icmlsetsymbol{equal}{*}

\begin{icmlauthorlist}
\icmlauthor{Shentong Mo}{cmu,mbz}
\icmlauthor{Haofan Wang}{xhs}
\end{icmlauthorlist}

\icmlaffiliation{cmu}{Carnegie Mellon University}
\icmlaffiliation{mbz}{MBZUAI}
\icmlaffiliation{xhs}{Xiaohongshu}

\icmlcorrespondingauthor{Shentong Mo}{shentongmo@gmail.com}

\icmlkeywords{Machine Learning, ICML}

\vskip 0.3in
]



\printAffiliationsAndNotice{}  

\begin{abstract}

Visual sound localization is a typical and challenging problem that predicts the location of objects corresponding to the sound source in a video.
Previous methods mainly used the audio-visual association between global audio and one-scale visual features to localize sounding objects in each image.
Despite their promising performance, they omitted multi-scale visual features of the corresponding image, and they cannot learn discriminative regions compared to ground truths.
To address this issue, we propose a novel multi-scale multi-instance visual sound localization framework, namely M2VSL, that can directly learn multi-scale semantic features associated with sound sources from the input image to localize sounding objects.
Specifically, our M2VSL leverages learnable multi-scale visual features to align audio-visual representations at multi-level locations of the corresponding image.
We also introduce a novel multi-scale multi-instance transformer to dynamically aggregate multi-scale cross-modal representations for visual sound localization.
We conduct extensive experiments on VGGSound-Instruments, VGG-Sound Sources, and AVSBench benchmarks.
The results demonstrate that the proposed M2VSL can achieve state-of-the-art performance on sounding object localization and segmentation.

\end{abstract}

\section{Introduction}

The ability to discern the location of a sound source in a visual context, much like our natural ability to identify the position of a barking dog within a room, has sparked significant interest in the field of audio-visual learning. This multidisciplinary field integrates audio signals with visual data, aiming to enhance our understanding of the environment around us. Our research focuses on the aspect of visual sound localization, where we aim to pinpoint the exact location of sound-producing objects within a visual frame.

Audio-visual learning poses unique challenges, as it requires the establishment of meaningful cross-modal correspondences between audio and visual inputs for a variety of tasks, including classification~\cite{pian2023audiovisual,mo2023classincremental}, localization~\cite{mo2022benchmarking,mo2022EZVSL,mo2022SLAVC,mo2023avsam,mo2023weakly}, and source separation~\cite{mo2024semantic}. Traditional methods in this field have employed various frameworks tailored to individual tasks, often isolating them from one another. For example, sound source localization (SSL) has been approached using two-stream neural networks and attention mechanisms, as seen in Attention10k, or through the lens of multiple-instance contrastive learning in EZVSL. However, these methods have often overlooked the multi-scale visual features of images and struggled with issues such as overfitting and silence in source sound localization.

\begin{figure}
\centering
\includegraphics[width=1.0\linewidth]{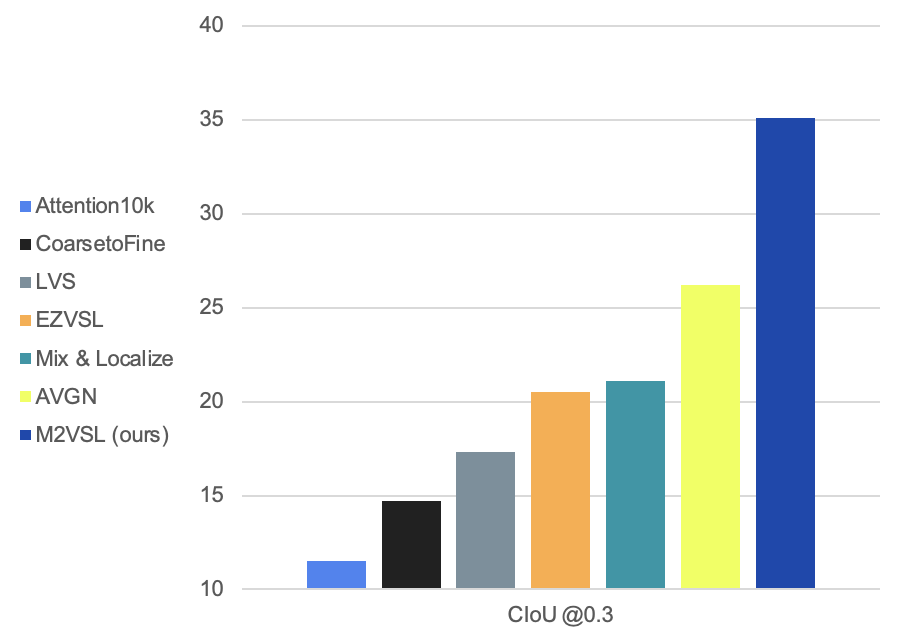}
\vspace{-1.5em}
\caption{{\bf Comparison of M2VSL with state-of-the-art methods on multi-source visual sound localization (Class-aware IoU@0.3).}}
\label{fig: title_img}
\end{figure}

A new and related challenge has emerged with audio-visual segmentation (AVS), which aims to generate pixel-level maps of sound-emitting objects. While SSL focuses on identifying rough locations of sound sources, AVS demands a more detailed and exacting approach, necessitating sophisticated network architectures. However, these intricate networks require extensive pixel-level annotations, which are resource-intensive to obtain. Our goal is to bypass the need for such detailed annotations by adopting a weakly-supervised approach to AVS.
One of the main hurdles in this new multi-modal, weakly-supervised problem is the absence of pixel-level segmentation masks during training. This gap significantly hampers the performance of AVS, as observed in our experiments. Additionally, existing weakly-supervised semantic segmentation and visual sound source localization methods face their own challenges, such as a lack of multi-modal constraints for accurate mask prediction and the production of only coarse heatmaps rather than detailed segmentation masks.

To overcome these challenges, we propose a novel approach: the Multi-scale Multi-instance Visual Sound Localization (M2VSL) framework. M2VSL is designed to align multi-scale visual features with audio signals across multiple levels. This approach diverges from previous methods by implementing multi-scale multiple-instance contrastive learning within the audio-visual fusion process. Furthermore, we introduce a unique component, the Multi-scale Multi-instance Transformer to dynamically aggregate multi-scale cross-modal representations, enhancing the standard Visual Transformer (ViT) specifically for the task of visual sound localization.

Through extensive experimentation on benchmarks such as VGGSound-Instruments, VGG-Sound Sources, and AVSBench, our M2VSL framework has demonstrated superior performance in localizing and segmenting sounding objects, setting new benchmarks in the field.

Our contributions are threefold:
\begin{itemize}
\item The development of the M2VSL framework, which uniquely aligns multi-scale visual features with audio signals at multiple levels.
\item The introduction of the  Multi-scale Multi-instance Transformer, a novel transformer model tailored for enhanced performance in visual sound localization.
\item Empirical evidence from extensive experiments showcasing the state-of-the-art performance of M2VSL in localizing and segmenting sounding objects.
\end{itemize}

\section{Related Work}

\subsection{Audio-Visual Learning}

In the realm of audio-visual learning, significant research~\citep{aytar2016soundnet,owens2016ambient,Arandjelovic2017look,korbar2018cooperative,Senocak2018learning,zhao2018the,zhao2019the,Gan2020music,Morgado2020learning,Morgado2021robust,Morgado2021audio,mo2023diffava,mo2023oneavm,mo2023deepavfusion,zhang2024audiosynchronized,mahmud2024maavt,mo2024audiovisual} has been directed towards establishing correlations between auditory and visual modalities within video sequences. 
The primary objective in these studies is to bridge the gap between disparate audio and visual pairs, enhancing their congruency while distinguishing them from mismatched pairs. 
This cross-modal correspondence has been instrumental in a variety of applications, including audio/speech separation~\citep{Gan2020music,Gao2018learning,Gao2019co,zhao2018the,zhao2019the,gao2020listen,tian2021cyclic,gao2021visualvoice}, audio spatialization~\citep{Morgado2018selfsupervised,gao20192.5D,Chen2020SoundSpacesAN,Morgado2020learning}, visual sound source localization~\citep{Senocak2018learning,Rouditchenko2019SelfsupervisedAC,hu2019deep,Afouras2020selfsupervised,qian2020multiple,chen2021localizing,mo2022EZVSL,mo2022SLAVC}, and audio-visual parsing~\citep{tian2020avvp,wu2021explore,lin2021exploring,mo2022multimodal,mo2022semantic}. 
Our research specifically zeroes in on visual sound localization and segmentation, representing a more intricate and demanding challenge than the tasks previously mentioned.

\subsection{Visual Sound Localization}

The evolution of visual sound source localization, identifying video regions corresponding to sound sources, has transitioned from early statistical models to sophisticated deep learning architectures. 
Early methods~\citep{hershey1999audio,fisher2000learning,kidron2005pixels} utilized canonical correlation analysis\citep{kidron2005pixels} , but recent developments\citep{Senocak2018learning,hu2019deep,Afouras2020selfsupervised,qian2020multiple,chen2021localizing,arda2022learning,mo2022EZVSL,mo2022SLAVC,mo2024texttoaudio,mo2024unified} have introduced deep neural networks for more nuanced audio-visual correspondence.
Notable contributions include two-stream architectures with attention mechanisms (e.g., Attention10k~\citep{Senocak2018learning}), and hard sample mining in LVS~\citep{chen2021localizing} for refined correspondence mapping. Further advancements have been seen in frameworks like EZVSL~\citep{mo2022EZVSL}, employing multiple-instance contrastive learning for improved region-sound alignment. 
Contemporary research~\citep{qian2020multiple,hu2020dsol,hu2022mix} has also tackled the complexity of localizing multiple sound sources in mixed audio environments, with approaches like silence-aware localization~\citep{hu2020dsol} and contrastive random walk models~\citep{hu2022mix} for audio-visual linking. 
Our approach, the Multi-scale Multi-instance Visual Sound Localization (M2VSL) framework, diverges from these methods by utilizing multi-scale visual features and a novel transformer model for enhanced localization and segmentation.

\begin{figure*}[t]
\centering
\includegraphics[width=0.9\linewidth]{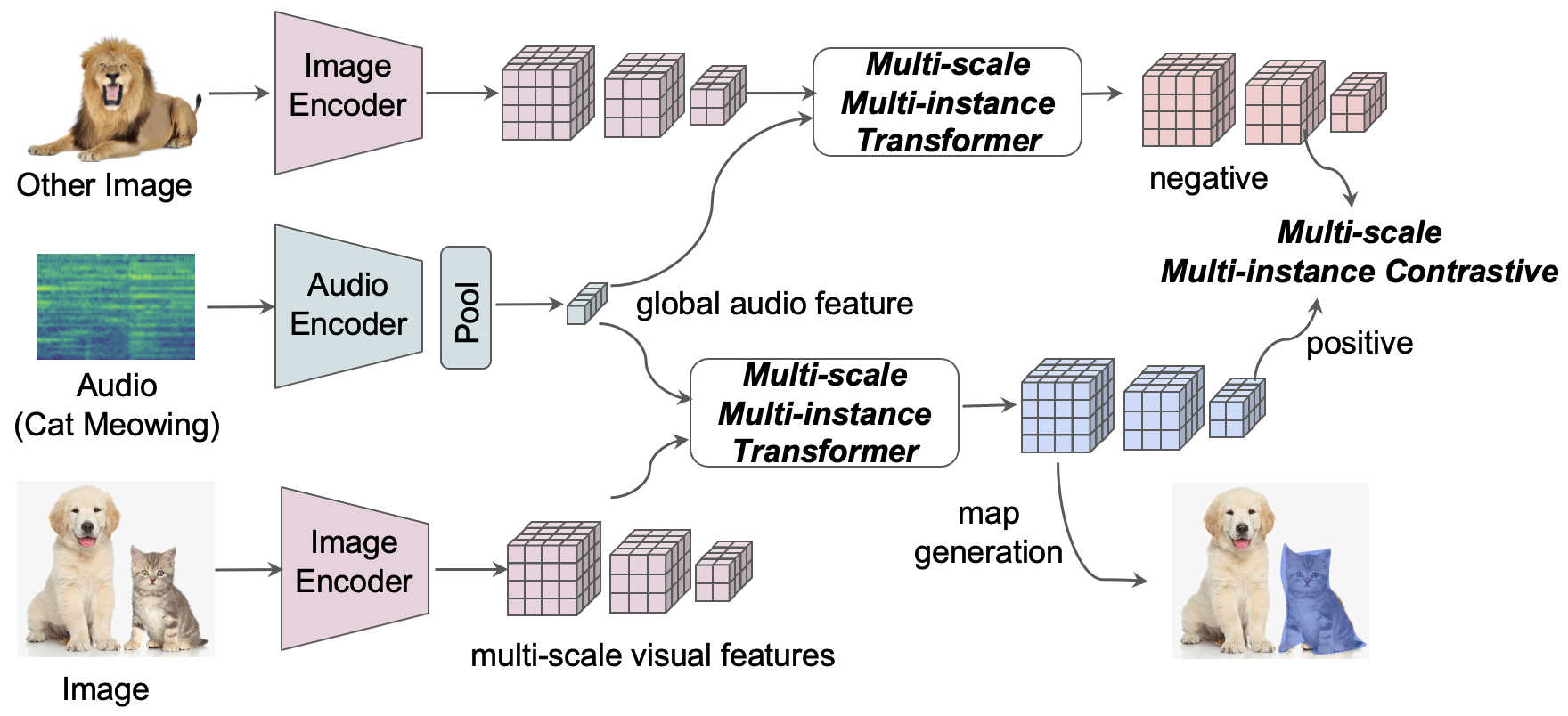}
\caption{{\bf Illustration of the proposed Multi-scale Multi-instance Visual Sound Localization (M2VSL) framework for weakly-supervised audio-visual localization and segmentation.}
}
\label{fig: main_img}
\end{figure*}

\subsection{Audio-Visual Segmentation}

Audio-visual segmentation, predicting pixel-level masks for sound-producing objects, remains a challenging task. This field was pioneered by Zhou et al.~\citep{zhou2022avs} with the introduction of a benchmark and an encoder-decoder network designed for pixel-level segmentation. 
The primary limitation of their method is its heavy reliance on extensive pixel-level annotations~\cite{}, leading to reduced effectiveness without such detailed supervision. 
In contrast, our work aims to overcome this limitation through a weakly-supervised approach, leveraging the M2VSL framework to achieve state-of-the-art performance in localizing and segmenting sounding objects without requiring exhaustive pixel-level annotations.

\section{Method}

In this paper, we introduce the Multi-scale Multi-instance Visual Sound Localization (M2VSL) framework, a novel approach for audio-visual localization and segmentation that eschews the need for pixel-wise annotation. M2VSL is composed of two primary modules: the Multi-scale Multiple Instance Contrastive (M$^2$IC) module and the Multi-scale Multi-instance Transformer (MMT) module.

\subsection{Preliminaries}

In this section, we first describe the problem setup and notations, and then revisit the multiple-instance contrastive learning in EZVSL~\citep{mo2022EZVSL} for single-source localization.

\noindent\textbf{Problem Setup and Notations.}

We consider the scenario where we are given a mixed spectrogram and an image, and our task is to spatially localize $N$ individual sound sources within the image. For a video containing $C$ source event categories, we are equipped with an audio-visual label, represented as ${y_i}_{i=1}^C$, where $y_i$ indicates the presence (1) or absence (0) of the ground-truth category $i$. It's important to note that during training, we lack bounding box or mask-level annotations and must rely solely on video-level labels for weakly-supervised learning.

\noindent\textbf{Revisit EZVSL.}

EZ-VSL introduced a multiple-instance contrastive learning framework for single-source localization by aligning spatial-level visual features with global audio features. This method utilizes a loss function that encourages the alignment of at least one visual feature location with the corresponding audio representation within the same mini-batch, defined as:
\begin{equation}\label{eq:micl}
    \mathcal{L}_{mc} = 
    - \log \frac{
    \exp \left( \frac{1}{\tau} \mathtt{sim}(\mathbf{A}, \mathbf{V}) \right)
    }{
    \sum_{k=1}^B \exp \left(  \frac{1}{\tau} \mathtt{sim}(\mathbf{A}, \mathbf{V}_k)\right)}
\end{equation}
The similarity metric, denoted as $\mathtt{sim}(\mathbf{A}, \mathbf{V})$, computes the max-pooled audio-visual cosine similarity across all spatial locations, leveraging batch size $B$ and a temperature hyper-parameter $\tau$.

\subsection{Multi-scale Multi-instance Contrastive}\label{sec:mmc}

Addressing the modality uncertainty inherent in previous weakly-supervised semantic segmentation baselines, our approach, inspired by EZ-VSL, focuses on aligning the audio with the most closely associated multi-scale visual features. This is predicated on the understanding that most video frame locations are unrelated to the sound source and should not be aligned with the audio during training.

We introduce a multi-scale multiple-instance contrastive learning objective, MMC, which seeks to align at least one location in a bag of multi-scale visual features with the corresponding audio representation in the same mini-batch, which is defined as:
\begin{equation}\label{eq:m2icl}
    \mathcal{L}_{a\rightarrow v} = 
    - \sum_{s=1}^S \log \frac{
    \exp \left( \frac{1}{\tau} \mathtt{sim}(\mathbf{A}, \mathbf{V}^s) \right)
    }{
    \sum_{k=1}^B \exp \left(  \frac{1}{\tau} \mathtt{sim}(\mathbf{A}, \mathbf{V}^s_k)\right)}
\end{equation}
The similarity metric, $\mathtt{sim}(\mathbf{A}, \mathbf{V}^s)$, computes the max-pooled audio-visual cosine similarity across all spatial locations at each scale stage.
 $\mathtt{sim}(\mathbf{A}, \mathbf{V}^s) = \max_{H^sW^s}(\mathbf{A}, \mathbf{V}^s)$.
Furthermore, we employ a symmetric loss to differentiate negative audio bags from other audio samples in the same mini-batch,
which is defined as 
\begin{equation}\label{eq:m2icl}
    \mathcal{L}_{v\rightarrow a} = 
    - \sum_{s=1}^S \log \frac{
    \exp \left( \frac{1}{\tau} \mathtt{sim}(\mathbf{A}, \mathbf{V}^s) \right)
    }{
    \sum_{k=1}^B \exp \left(  \frac{1}{\tau} \mathtt{sim}(\mathbf{A}_k, \mathbf{V}^s)\right)}
\end{equation}
where $\mathbf{A}_k$ denote the global audio visual from other sample $k$ in the mini-batch.
The overall audio-visual fusion objective, leveraging the MMC mechanism, is given as:
\begin{equation}
\begin{aligned}
    \mathcal{L}_{mmc} = - \sum_{s=1}^S \log \frac{
    \exp \left( \frac{1}{\tau} \mathtt{sim}(\mathbf{A}, \mathbf{V}^s) \right)
    }{
    \sum_{k=1}^B \exp \left(  \frac{1}{\tau} \mathtt{sim}(\mathbf{A}, \mathbf{V}^s_k)\right)} + \\
    \log \frac{
    \exp \left( \frac{1}{\tau} \mathtt{sim}(\mathbf{A}, \mathbf{V}^s) \right)
    }{
    \sum_{k=1}^B \exp \left(  \frac{1}{\tau} \mathtt{sim}(\mathbf{A}_k, \mathbf{V}^s)\right)}
\end{aligned}
\end{equation}

This approach aims to learn discriminative global audio representations and multi-scale visual features, which are then used to generate updated multi-scale audio-visual features and, ultimately, the output mask using $\mathtt{sim}(\mathbf{A}, \mathbf{V}^s) = \max_{H^sW^s}(\mathbf{A}, \mathbf{V}^s)$, which follows EZ-VSL~\cite{mo2022EZVSL}.

\subsection{Multi-scale Multi-instance Transformer}\label{sec:mmt}

The MMT module is designed to effectively aggregate multi-scale features from the raw input. 
Using self-attention transformers $\phi(\cdot)$, we process the categorical token embeddings $\{\hat{\mathbf{c}}_i^{av}\}_{i=1}^C$ to update feature representations, as defined as:
\begin{equation}
\begin{aligned}
    \{\hat{\mathbf{f}}^{av}_{p}\}_{p=1}^P, \{\hat{\mathbf{c}}_i\}_{i=1}^C = \{\phi(\mathbf{x}_j^{av}, \mathbf{X}^{av}, \mathbf{X}^{av})\}_{j=1}^{P+C}, \\
    \mathbf{X}^{av} = \{\mathbf{x}_j^{av}\}_{j=1}^{P+C} = [\{\mathbf{f}_{p}^{av}\}_{p=1}^P; \{\mathbf{c}_i^{av}\}_{i=1}^C]
\end{aligned}
\end{equation}
where $[\ ;\ ]$ denotes the concatenation operator.
$\mathbf{f}_p^{av},\mathbf{c}_i^{av},\mathbf{x}_j^{av}\in\mathbb{R}^{1\times D}$, and $D$ is the dimension of embeddings. 
The self-attention operator $\phi(\cdot)$ is formally defined as:
\begin{equation}
    \phi(\mathbf{x}_j^{av}, \mathbf{X}^{av}, \mathbf{X}^{av}) = \mbox{Softmax}(\dfrac{\mathbf{x}_j^{av}(\mathbf{X}^{av})^\top}{\sqrt{D}})\mathbf{X}^{av},
\end{equation}
where the softmax function is applied to the scaled dot-product of the input embeddings, facilitating the aggregation of relevant information across different feature embeddings.

Overall, the M2VSL framework, with its MMC and MMT modules, presents a robust solution for audio-visual localization and segmentation, effectively bypassing the need for extensive pixel-level annotations and setting new standards for performance in the field.

\begin{table*}[t]
	\renewcommand\tabcolsep{6.0pt}
	\centering
	\scalebox{0.9}{
		\begin{tabular}{l|ccc|cccc}
			\toprule
			\multirow{2}{*}{Method} & \multicolumn{3}{c|}{VGGSound-Instruments} & \multicolumn{3}{c}{VGGSound-Single}  \\
			&  AP(\%) & IoU@0.3(\%) & AUC(\%) &  AP(\%) & IoU@0.5(\%) & AUC(\%) \\ 	
			\midrule
			Attention10k~\cite{Senocak2018learning} & -- & 28.3 & 26.1 & -- & 19.2 & 30.6 \\
               OTS~\cite{Arandjelovic2018ots} & 47.5 & 25.7 & 24.6 & 29.8 & 32.8 & 35.7 \\

		   DMC~\cite{hu2019deep} & -- & 26.5 & 25.7 & -- & 23.9 & 27.6 \\
            CoarsetoFine~\cite{qian2020multiple} & 40.2 & 27.2 & 26.5 & 28.2 & 29.1 & 34.8 \\

			LVS~\cite{chen2021localizing}  & 42.3 & 32.6 & 28.3 & 29.6 & 34.4 & 38.2 \\

			EZ-VSL~\cite{mo2022EZVSL}  & 43.8 & 38.5 & 30.6 & 31.3 & 38.9 & 39.5 \\

                Mix-and-Localize~\cite{hu2022mix}  & 44.9 & 49.7 & 32.3 & 32.5 & 36.3 & 38.9 \\

                DSOL~\cite{hu2020dsol}  & -- & 50.2 & 32.9 & -- & 35.7 & 37.2 \\

                AVGN~\cite{mo2023audiovisual} & 50.5 & 55.3 & 36.7 & 35.3 & 40.8 & 42.3 \\

                M2VSL (ours) & \bf 53.9 & \bf 59.7 & \bf 39.8 & \bf 40.6 & \bf 46.8 & \bf 50.2 \\

			\bottomrule
			\end{tabular}}
   \caption{{\bf Quantitative results of single-source localization on VGGSound-Instruments and VGGSound-Single datasets.}}
   \label{tab: exp_sota_single}
\end{table*}

\begin{table*}[t]
	\renewcommand\tabcolsep{4.0pt}
	\centering
	\scalebox{0.8}{
		\begin{tabular}{l|cccc|ccccc}
			\toprule
			\multirow{2}{*}{Method} & \multicolumn{4}{c|}{VGGSound-Instruments} &  \multicolumn{4}{c}{VGGSound-Duet}  \\
			& CAP(\%) & PIAP(\%) & CIoU@0.1(\%) & AUC(\%) & CAP(\%) & PIAP(\%) & CIoU@0.3(\%) & AUC(\%) \\ 	
			\midrule
			Attention10k~\cite{Senocak2018learning} & -- & -- & 52.3 & 11.7 & -- & -- & 11.5 & 15.2 \\

            OTS~\cite{Arandjelovic2018ots} & 23.3 & 37.8 & 51.2 & 11.2 & 10.5 & 12.7 & 12.2 & 15.8 \\

			DMC~\cite{hu2019deep} & -- & -- & 53.7 & 12.5 & -- & -- & 13.8 & 17.1 \\

   CoarsetoFine~\cite{qian2020multiple} & -- & -- & 54.2 & 12.9 & -- & -- & 14.7 & 18.5 \\

			LVS~\cite{chen2021localizing} & -- & -- & 57.3 & 13.3 & -- & -- & 17.3 & 19.5 \\

			EZ-VSL~\cite{mo2022EZVSL}  & -- & -- & 60.2 & 14.2 & -- & -- & 20.5 & 20.2 \\

                Mix-and-Localize~\cite{hu2022mix}  & 21.5 & 37.5 & 73.2 & 15.6 & 16.3 & 22.6 & 21.1 & 20.5 \\

                DSOL~\cite{hu2020dsol} & -- & -- & 74.3 & 15.9 & -- & -- & 22.3 & 21.1 \\

                AVGN~\cite{mo2023audiovisual} & 27.3 & 42.8 & 77.5 & 18.2 & 21.9 & 28.1 & 26.2 & 23.8 \\
                
                M2VSL (ours) & \bf 32.1 & \bf 48.5 & \bf 82.3 & \bf 24.5 & \bf 28.3 & \bf 36.2 & \bf 35.1 & \bf 32.6 \\

			\bottomrule
			\end{tabular}}
   \caption{{\bf Quantitative results of multi-source localization on VGGSound-Instruments and VGGSound-Duet datasets.}}
   \label{tab: exp_sota_multi}
   
\end{table*}

\vspace{-0.5em}
\section{Experiments}
\vspace{-0.5em}

\subsection{Experimental setup}

\noindent\textbf{Datasets.}
VGGSound-Instruments, derived from~\cite{hu2022mix}, is a curated collection of 32,000 video clips, each 10 seconds in length, from 37 musical instrument categories. This dataset represents a subset of the larger VGG-Sound dataset ~\cite{chen2020vggsound} and is uniquely characterized by each video carrying a label for only a single instrument category. For multi-source localization assessment, the protocol follows that of~\cite{hu2022mix}, where two frames are randomly concatenated to form a single input image of dimensions $448\times 224$, and their respective waveforms are combined to create an audio mixture.

Additionally, our research extends beyond these musical datasets by incorporating 150,000 video clips, each 10 seconds long, from the original VGG-Sound dataset~\cite{chen2020vggsound}. This subset, termed VGGSound-Single, spans 221 diverse categories including nature scenes, animals, vehicles, people, and various instruments. The single-source localization testing employs the complete VGG-Sound Source test set, featuring 5,158 videos~\cite{chen2021localizing}. 
For multi-source localization, we adopt an approach similar to that used for VGGSound-Instruments, resulting in 5,158 mixed videos. This test set, referred to as VGGSound-Duet, presents a more complex challenge than the 446 videos in VGGSound-Instruments.

Furthermore, our study incorporates AVSBench~\cite{zhou2022avs}, consisting of 4,932 videos with a total of 10,852 frames from 23 categories, including animals, humans, and instruments. In alignment with~\cite{zhou2022avs}, we utilize a split of 3,452/740/740 videos for the train/validation/test segments in single-source segmentation.

\noindent\textbf{Evaluation Metrics.}
In alignment with the methodologies outlined in~\citep{hu2022mix}, our evaluation metrics for single-source localization encompass the average precision at the pixel-wise average precision (AP), Intersection over Union (IoU), and Area Under Curve (AUC). For multi-source localization assessments, we adopt the class-aware average precision (CAP), permutation-invariant average precision (PIAP), Class-aware IoU (CIoU), and Area Under Curve (AUC), ensuring a fair comparison with the standards set in~\citep{hu2022mix}.

The threshold settings for IoU and CIoU vary based on the dataset. Specifically, for both single-source and multi-source localization on VGGSound-Instruments, the thresholds are set at IoU@0.3 and CIoU@0.1. Similarly, for single-source and multi-source localization on VGGSound-Single and VGGSound-Duet, we employ IoU@0.5 and CIoU@0.3.
Consistent with established practices in prior research~\cite{zhou2022avs}, the evaluation of audio-visual segmentation performance is conducted using the averaged IoU (mIoU) and the F-score. The mIoU metric is designed to compute the intersection-over-union between the predicted mask and the ground-truth mask, serving as a measure of region similarity. 
The F-score, on the other hand, is calculated to assess contour accuracy by considering both precision and recall.

\noindent\textbf{Implementation.}
In our experimental setup, the input image resolution is adjusted to $224 \times 224$ pixels. For the audio input, we extract log spectrograms from 3-second audio clips at a sampling rate of 22,050 Hz. Consistent with the approach in~\cite{mo2022EZVSL}, we employ the Short-Time Fourier Transform (STFT) to generate an input tensor of dimensions $257 \times 300$, corresponding to $257$ frequency bands over $300$ time steps. This process utilizes a window size of 50 milliseconds and a hop size of 25 milliseconds.
In line with methodologies adopted in previous studies by~\citep{hu2019deep,qian2020multiple,chen2021localizing,mo2022EZVSL,mo2022SLAVC}, we select the lightweight ResNet18 architecture as our audio and visual encoder, as proposed by~\citep{he2016resnet}. 
The visual model is initialized with weights pre-trained on the ImageNet dataset~\citep{imagenet_cvpr09}. The chosen dimension size for our model is $D=512$, and we utilize $P=25$ for the $5\times 5$ spatial map output from $S=4$ stages by the visual encoder.
The self-attention transformers in our model, denoted as $\phi^a(\cdot)$ and $\phi^v(\cdot)$, are set to a depth of 3 layers. 
The training regimen for the model encompasses 100 epochs, utilizing the Adam optimizer with a learning rate of $1e-4$ and a batch size of 128, as per the guidelines established by~\citep{kingma2014adam}.

\begin{figure*}[t]
\centering
\includegraphics[width=0.9\linewidth]{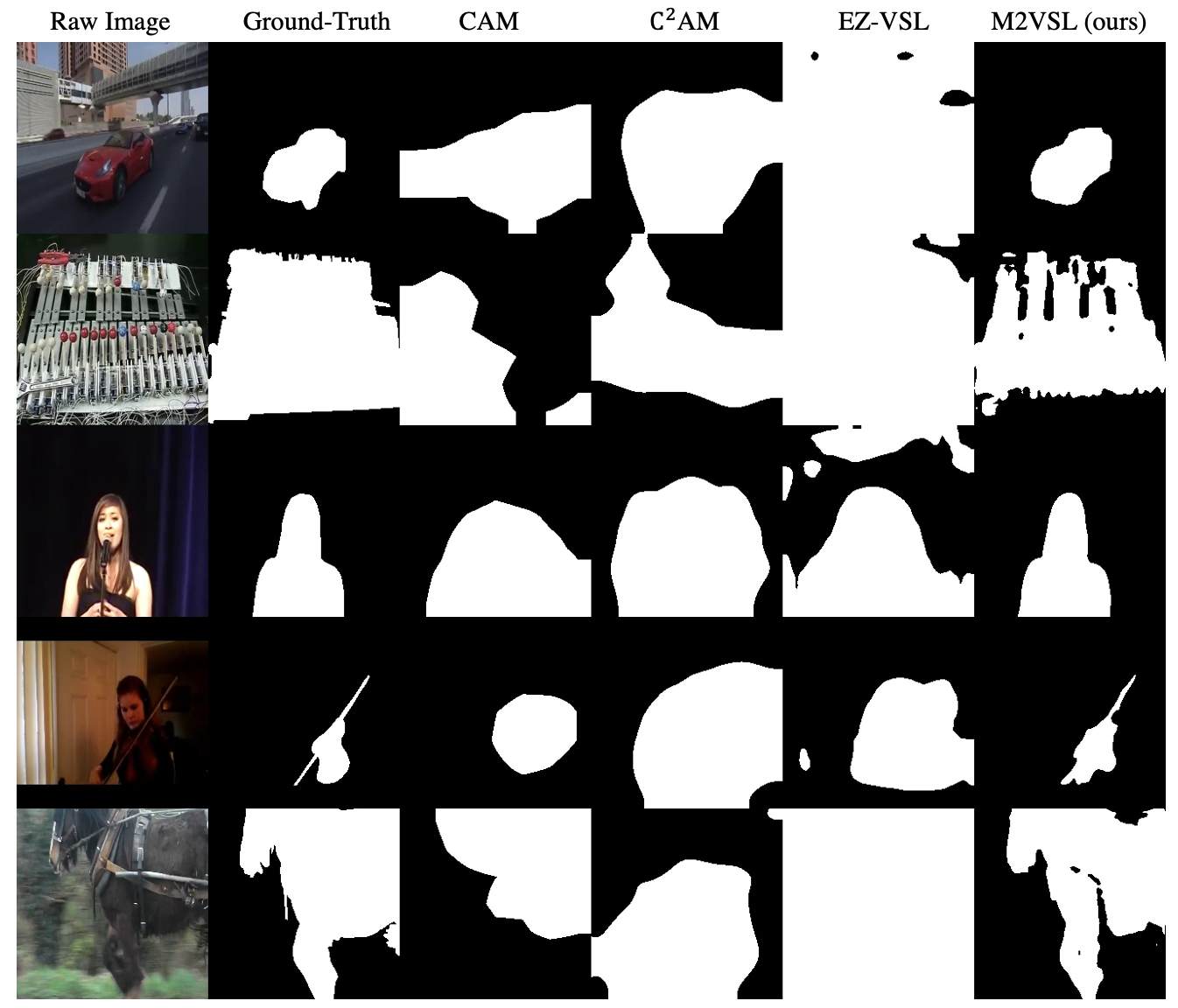}
\vspace{-0.6em}
\caption{{\bf Qualitative comparisons with weakly-supervised semantic segmentation and visual sound source localization baselines.} 
The proposed M2VSL generates more accurate and high-quality segmentation maps for sounding objects.
}
\label{fig: exp_vis}
\end{figure*}

\vspace{-0.5em}
\subsection{Comparison to prior work}\label{sec:exp}
\vspace{-0.5em}

In our study, we introduce the Multi-scale Multi-instance Visual Sound Localization (M2VSL) framework, a novel and effective approach for sound source localization and segmentation. To demonstrate the efficacy of M2VSL, we conduct a comprehensive comparison with established single-source and multi-source baselines in the field:
1) Attention 10k~\citep{Senocak2018learning}  (2018'CVPR): pioneering the field of single-source localization, this work utilizes a two-stream architecture combined with an attention mechanism, setting the groundwork for subsequent studies in this area;
2) OTS~\citep{Arandjelovic2018ots} (2018'ECCV): a baseline approach that focuses on learning audio-visual correspondence, offering a straightforward yet effective method for sound source localization;
3) DMC~\citep{hu2019deep} (2019'CVPR): this deep multi-modal clustering network leverages audio-visual co-occurrences to learn convolutional maps for each modality in distinct embedding spaces, enhancing the ability to discern multimodal data;
4) CoarsetoFine~\citep{qian2020multiple} (2020'ECCV): a two-stage method that employs a coarse-to-fine strategy for aligning cross-modal features, aiming to improve the precision of localization;
5) DSOL~\citep{hu2020dsol} (2020'NeurIPS): this framework adopts a two-stage training approach, utilizing class labels as weak supervision for category-aware sound source localization;
6) LVS~\citep{chen2021localizing} (2021'CVPR): a contrastive network designed to learn audio-visual correspondence maps, incorporating hard negative mining to refine its localization capabilities;
7) EZ-VSL~\citep{mo2022EZVSL} (2022'ECCV): a recent strong baseline that implements multiple-instance contrastive learning specifically for single-source localization;
8) Mix-and-Localize~\citep{hu2022mix} (2022'CVPR): this robust multi-source baseline utilizes a contrastive random walk algorithm within a graph composed of images and separated sounds as nodes, offering a novel approach to multi-source localization;
9) AVGN~\cite{mo2023audiovisual} (2023'CVPR): a cutting-edge grouping-based method that disentangles category-wise semantic features from both the audio mixture and image, enabling simultaneous localization of multiple sounding sources.

\begin{table*}[t]
	\renewcommand\tabcolsep{10.0pt}
    \renewcommand{\arraystretch}{1.2}
	\centering
	\scalebox{0.95}{
		\begin{tabular}{lcccc}
		\toprule
		\multirow{2}{*}{\bf Method} & \multicolumn{2}{c}{\bf Single Source} & \multicolumn{2}{c}{\bf Multiple Source} \\
            & \bf mIoU & \bf F-score  & \bf mIoU & \bf F-score  \\
		\midrule
		AVS~\cite{zhou2022avs} (ws) & 12.63 & 24.99 & 8.76	& 15.72 \\
            CAM~\cite{zhou2016learning} & 19.26 & 27.88 & 12.65	& 19.83 \\
            EZ-VSL~\cite{mo2022EZVSL} & 29.40 & 35.70 & 23.58	& 27.31 \\
            C$^2$AM~\cite{xie2022c2am} & 30.87 & 36.55 & 25.33	& 29.58 \\
            M2VSL (ours) & \bf 37.85 & \bf 55.21 & \bf 35.26  & \bf 49.35  \\
		\bottomrule
			\end{tabular}}
   \caption{{\bf Comparison results (\%) of weakly-supervised audio-visual segmentation.} ``ws'' refers to the weakly-supervised, where only the instance-level category label is used during the training stage.}
	\label{tab: exp_sota_seg}
\end{table*}

\noindent\textbf{Single-source sound localization.}
In the realm of single-source localization, our quantitative results, detailed in Table~\ref{tab: exp_sota_single}, exhibit M2VSL's superior performance across all evaluated metrics on two benchmark datasets, surpassing both self-supervised and weakly-supervised baselines. 
Notably, M2VSL achieves remarkable improvements over AVGN~\citep{mo2023audiovisual}, the current state-of-the-art, enhancing the IoU@0.3 and AUC by 4.4 and 3.1 points, respectively, on VGGSound-Instruments, and enhancing IoU@0.5 and AUC by 6.0 and 7.9 points, respectively, on VGGSound-Single.
These advancements are particularly significant when compared to EZ-VSL~\citep{mo2022EZVSL}, a strong contender in the self-supervised domain. This comparison underscores the critical role of multi-scale semantic extraction from multiple audio-visual instances in learning discriminative audio-visual alignment. Additionally, M2VSL exhibits substantial outperformance over Mix-and-Localize~\citep{hu2022mix}, achieving gains of 9.0 in AP on VGGSound-Instruments and 8.1 in AP on VGGSound-Single.
Such pronounced improvements firmly establish the efficacy of our M2VSL framework in the field of single-source localization, demonstrating its potential to set new standards in audio-visual learning.

\noindent\textbf{Multi-source sound localization.}
Our findings, as detailed in Table~\ref{tab: exp_sota_multi}, also indicate substantial improvements in multi-source sound localization with the M2VSL framework. In comparison to AVGN~\citep{mo2023audiovisual}, the leading multi-source localization benchmark, M2VSL demonstrates significant enhancements, achieving gains of 4.8 in CAP, 5.7 in PIAP, 4.8 in CIoU@0.1, and 6.3 in AUC on the VGGSound-Instruments dataset.
When assessed on the more demanding VGGSound-Duet benchmark, M2VSL continues to surpass AVGN, with notable improvements of 6.4 in CAP, 8.1 in PIAP, 8.9 in CIoU@0.3, and 8.8 in AUC. This outperformance is not just limited to comparisons with AVGN but extends to DSOL~\citep{hu2020dsol}, a robust weakly-supervised baseline known for its two-stage training process.
These results underscore the effectiveness of M2VSL in deciphering and learning multi-scale source semantics from both audio mixtures and corresponding images, thus bolstering its capabilities in multi-source sound localization. 
Such achievements highlight M2VSL's potential as a groundbreaking framework in the realm of audio-visual learning.

\noindent\textbf{Single-source audio-visual segmentation.}
The quantitative results for audio-visual segmentation are presented in Table~\ref{tab: exp_sota_seg}, illustrating the superior performance of M2VSL across all metrics when compared to existing weakly-supervised baselines. Notably, M2VSL substantially surpasses the weakly-supervised version of the state-of-the-art audio-visual segmentation approach, AVS~\cite{zhou2022avs}, with remarkable improvements of 25.22 in mIoU and 30.31 in F-score.
Furthermore, M2VSL demonstrates its efficacy by achieving significant performance gains over C$^2$AM~\cite{xie2022c2am}, with an increase of 6.98 in mIoU and 18.66 in F-score. This comparison underscores the critical contribution of our multi-scale multi-instance contrastive learning and transformer architecture in enhancing audio-visual segmentation capabilities.
Additionally, when compared to EZ-VSL (Mo et al., 2022), a strong baseline in visual sound source localization, M2VSL shows substantial superiority, recording performance gains of 8.45 in mIoU and 19.51 in F-score. These considerable advancements firmly establish M2VSL's leading position in the field of single-source audio-visual segmentation, highlighting its potential to set new benchmarks in this domain.

\noindent\textbf{Multi-source audio-visual segmentation.}
The results for multi-source audio-visual segmentation, as detailed in Table~\ref{tab: exp_sota_seg}, demonstrate notable enhancements achieved by our M2VSL framework. Compared to the weakly-supervised version of the state-of-the-art audio-visual segmentation method, AVS~\cite{zhou2022avs}, M2VSL registers substantial gains of 26.50 in mIoU and 33.63 in F1 score.
Moreover, M2VSL significantly outperforms C$^2$AM~\cite{xie2022c2am}, a contemporary approach in the field, with an improvement of 9.93 in mIoU and 19.77 in F1 score. This advancement not only highlights the efficacy of M2VSL but also underlines the importance of our multi-scale multi-instance approach in audio-visual segmentation.
Additionally, M2VSL's performance surpasses that of EZ-VSL~\cite{mo2022EZVSL}, a strong baseline in visual sound source localization. The comparative results affirm the superiority of M2VSL in handling the complexities of multi-source audio-visual segmentation, showcasing its capability to effectively learn and leverage multi-scale multi-instance semantics from both audio mixtures and corresponding images.
Overall, these results solidify M2VSL's position as an effective solution for multi-source audio-visual segmentation, validating its innovative approach in the realm of multi-modal learning.

\noindent\textbf{Qualitative visualizations.}
In our qualitative assessment of localization maps, we juxtapose the performance of M2VSL with that of other notable methodologies, including CAM~\cite{zhou2016learning}, C$^2$AM~\cite{xie2022c2am}, EZ-VSL~\citep{mo2022EZVSL}. 
This comparative analysis, illustrated in Figure~\ref{fig: exp_vis}, yields several key observations:
As a weakly-supervised object localization baseline, CAM demonstrates limitations in multi-source localization scenarios. Lacking explicit separation objectives, it tends to underperform in complex multi-source environments.
When compared to EZ-VSL, a self-supervised audio-visual baseline, our M2VSL framework exhibits a notably higher quality of localization maps. This improvement highlights the enhanced ability of M2VSL to accurately localize sound sources in a visual context.
Against the backdrop of the weakly-supervised multi-source baseline, C$^2$AM, M2VSL not only competes but in some cases, surpasses its performance in terms of the precision of predicted maps. This is particularly noteworthy considering that M2VSL benefits from the use of category labels during training, which aids in refining its localization capabilities.
These visual comparisons further underline the efficacy of M2VSL in learning and utilizing multi-scale audio-visual representations. This capability significantly contributes to the precise localization of each source, thereby reinforcing the superiority of M2VSL in the domain of audio-visual segmentation and localization.

\begin{table}[t]
	\renewcommand\tabcolsep{10.0pt}
    \renewcommand{\arraystretch}{1.2}
	\centering
	\scalebox{0.83}{
		\begin{tabular}{cccccc}
			\toprule
 \multirow{2}{*}{\bf MMC} & \multirow{2}{*}{\bf MMT} & \multicolumn{2}{c}{\bf Single Source} & \multicolumn{2}{c}{\bf Multiple Source} \\
            & & \bf mIoU & \bf F-score  & \bf mIoU & \bf F-score  \\
			\midrule
			\xmark & \xmark & 12.63 & 24.99 & 8.76	& 15.72 \\
               \cmark & \xmark & 34.87 & 51.95 & 30.92 & 47.05 \\
               \xmark & \cmark & 35.25 & 53.02 & 31.28 & 48.13 \\
               \cmark & \cmark & \bf 37.85 & \bf 55.21 & \bf 35.26 & \bf 49.35 \\
			\bottomrule
			\end{tabular}}
   \caption{{\bf Ablation studies on Multi-scale Multi-instance Contrastive (MMC) and Multi-scale Multi-instance Transformer (MMT).} }
	\label{tab: ab_module}
\end{table}

\subsection{Experimental Analysis}

In this section, we conduct ablation studies to ascertain the contributions of the Multi-scale Multi-instance Contrastive (MMC) and the Multi-scale Multi-instance Transformer (MMT) modules within our M2VSL framework. Additionally, we explore the impact of batch size on weakly-supervised audio-visual segmentation.

\noindent{\textbf{Multi-scale Multi-instance Con- trastive \& Multi-scale Multi-instance Transformer.}}
To evaluate the effectiveness of integrating MMC and MMT for audio-visual fusion, we perform experiments to assess the necessity and impact of each module separately and in combination. The quantitative results of these ablations are presented in Table~\ref{tab: ab_module}. 
We observe notable improvements in single-source audio-visual segmentation upon adding MMC to the baseline model, with increases of 22.24 in mIoU and 26.96 in F-score. This enhancement highlights the efficacy of the MMC in extracting well-aligned cross-modal features crucial for source segmentation.
Similarly, the inclusion of only the MMT module in the baseline also results in improved segmentation performance across all evaluated metrics. Most notably, the combined integration of both MMC and MMT into the baseline leads to a substantial uplift in performance, achieving gains of 25.22 in mIoU and 30.22 in F-score. 
These results underscore the significance of both MMC and MMT modules in our M2VSL framework, confirming their critical role in addressing modality and spatial challenges effectively and producing precise audio-visual segmentation masks.

\begin{figure}[t]
\centering
\includegraphics[width=0.8\linewidth]{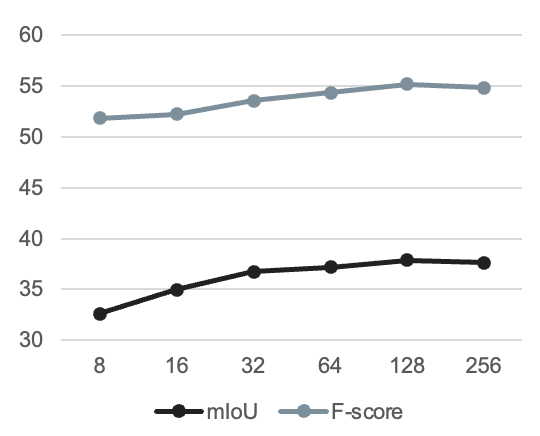}
\caption{{\bf Effect of batch size on eakly-supervised audio-visual segmentation (mIoU and F-score are reported)}.
}
\label{fig: ab_batch}
\end{figure}

\noindent{\textbf{Effect of Batch Size.}}
In our M2VSL framework, the batch size utilized in the Multi-scale Multi-instance Contrastive (MMC) plays a pivotal role in shaping the cross-modal representations crucial for audio-visual segmentation. To investigate this aspect thoroughly, we conducted experiments with varying batch sizes, specifically $\{8, 16, 32, 64, 128, 256\}$. The outcomes of these experiments, in terms of segmentation performance, are depicted in Figure~\ref{fig: ab_batch}.
Our findings indicate that a batch size of 128 in MMC yields the most optimal segmentation performance across all metrics. This result underscores the significance of an appropriately sized batch in the extraction and learning of discriminative cross-modal representations, particularly when leveraging multi-scale visual features in MMC.

\section{Conclusion}

In this work, we present M2VSL, a novel multi-scale multi-instance visual sound localization framework, that can directly learn multi-scale semantic features associated with sound sources from the input image to localize sounding objects.
Specifically, our M2VSL leverages learnable multi-scale visual features to align audio-visual representations at multi-level locations of the corresponding image.
We also introduce a novel multi-scale multi-instance transformer to dynamically aggregate multi-scale features for visual sound localization.
We conduct extensive experiments on VGGSound-Instruments, VGG-Sound Sources, and AVSBench benchmarks.
The results demonstrate that the proposed M2VSL can achieve state-of-the-art performance on sounding object localization and segmentation.

\section*{Impact Statement}

This research on Multi-scale Multi-instance Visual Sound Localization (M2VSL) presents significant implications and potential impacts across various sectors and societal aspects. Understanding and analyzing the broader impact of this work involves considering both the positive advancements it can foster and the potential challenges or ethical considerations it might raise.
M2VSL can significantly improve the user experience in multimedia applications. For instance, in virtual reality (VR) and augmented reality (AR), more accurate audio-visual localization can lead to immersive and realistic environments. 
This technique could also enhance the viewing experience in film and television by aligning sound and visuals more precisely.
This framework has the potential to aid in the development of assistive devices for individuals with sensory impairments. For the hearing impaired, more accurate visual localization of sound sources could lead to better speech recognition and understanding in complex environments.
In surveillance, the ability to accurately localize sound sources visually can be crucial. It can aid in identifying the source of distress calls or unusual noises, thereby enhancing public safety and security measures.
The techniques developed in M2VSL can be beneficial in scientific research areas like animal behavior studies, where understanding the correlation between visual cues and sounds is essential.

\bibliography{reference}
\bibliographystyle{icml2024}


\end{document}